\documentclass[sigconf]{acmart}

\copyrightyear{2020}
\acmYear{2020}
\setcopyright{acmcopyright}\acmConference[MM '20]{Proceedings of the 28th ACM International Conference on Multimedia}{October 12--16, 2020}{Seattle, WA, USA}
\acmBooktitle{Proceedings of the 28th ACM International Conference on Multimedia (MM '20), October 12--16, 2020, Seattle, WA, USA}
\acmPrice{15.00}
\acmDOI{10.1145/3394171.3413960}
\acmISBN{978-1-4503-7988-5/20/10}

\usepackage{array}
\newcommand{\myPara}[1]{\vspace{.05in}\noindent\textbf{#1}}
\begin{document}
\fancyhead{}
%%
%% The "title" command has an optional parameter,
%% allowing the author to define a "short title" to be used in page headers.
\title{Look Through Masks: Towards Masked Face Recognition with De-Occlusion Distillation}

%%
%% The "author" command and its associated commands are used to define
%% the authors and their affiliations.
%% Of note is the shared affiliation of the first two authors, and the
%% "authornote" and "authornotemark" commands
%% used to denote shared contribution to the research.
%%\author{Anonymous Submission}

\author{Chenyu Li}
\affiliation{%
  \institution{Institute of Information Engineering, Chinese Academy of Sciences, Beijing 100095, China \\School of Cyber Security, University of Chinese Academy of Sciences, Beijing 100049, China}
  %\streetaddress{1 Minzhuang}
  %\city{Beijing}
  %\country{Iceland}
  }
%\email{lichenyu@iie.ac.cn}

\author{Shiming Ge}\authornote{Shiming Ge is the corresponding author (geshiming@iie.ac.cn).}
%\authornote{Both authors contributed equally to this research.}
\affiliation{%
  \institution{Institute of Information Engineering, Chinese Academy of Sciences, Beijing 100095, China \\School of Cyber Security, University of Chinese Academy of Sciences, Beijing 100049, China}
  %\streetaddress{P.O. Box 1212}
  %\city{Beijing}
  %\country{China}
  %\postcode{100095}
}

\author{Daichi Zhang}
\affiliation{%
  \institution{Institute of Information Engineering, Chinese Academy of Sciences, Beijing 100095, China \\School of Cyber Security, University of Chinese Academy of Sciences, Beijing 100049, China}
  %\city{Beijing}
  %\country{China}
  }

\author{Jia Li}
\affiliation{
  \institution{State Key Laboratory of Virtual Reality Technology and Systems, SCSE, Beihang University, Beijing 100191, China \\Peng Cheng Laboratory, Shenzhen 518055, China}
  %\city{Beijing}
  }
%\email{jiali@buaa.edu.cn}

%%
%% By default, the full list of authors will be used in the page
%% headers. Often, this list is too long, and will overlap
%% other information printed in the page headers. This command allows
%% the author to define a more concise list
%% of authors' names for this purpose.
\renewcommand{\shortauthors}{C. Li, et al.}

%%
%% The abstract is a short summary of the work to be presented in the
%% article.

\begin{abstract}
  Many real-world applications today like video surveillance and urban governance need to address the recognition of masked faces, where content replacement by diverse masks often brings in incomplete appearance and ambiguous representation, leading to a sharp drop in accuracy. Inspired by recent progress on amodal perception, we propose to migrate the mechanism of amodal completion for the task of masked face recognition with an end-to-end de-occlusion distillation framework, which consists of two modules. The \textit{de-occlusion} module applies a generative adversarial network to perform face completion, which recovers the content under the mask and eliminates appearance ambiguity. The \textit{distillation} module takes a pre-trained general face recognition model as the teacher and transfers its knowledge to train a student for completed faces using massive online synthesized face pairs. Especially, the teacher knowledge is represented with structural relations among instances in multiple orders, which serves as a posterior regularization to enable the adaptation. In this way, the knowledge can be fully distilled and transferred to identify masked faces. Experiments on synthetic and realistic datasets show the efficacy of the proposed approach.

  %The question is how to represent these masked parts of perceived objects: this is the problem of amodal perception. Recent researches have proved importance of amodal completion in perceiving partly-observed objects.

  %The recognition of partially-occluded faces can facilitate many applications like video surveillance and urban governance, while it is challenged by the incomplete appearance and insufficient representation introduced by diverse occlusions. However, human beings have the ability to infer the missing content and identity of an occluded face by using the available content and their rich experience. Inspired by that, we propose a generation-to-adapt approach to migrate this ability by de-occlusion distillation in an end-to-end manner, which consists of two main modules. The de-occlusion module applies a generative adversarial network to perform face completion, which generates the missing part and eliminates the knowledge ambiguity of diverse occlusions. The distillation module takes a pre-trained normal face recognition model as teacher and transfer its knowledge to learn a robust representation model for completed faces from massive online synthesized face pairs. Especially, we extract various order structured relation knowledge among face instances. In this way, the knowledge can be fully distilled and effectively adapted to identify the occluded faces. Experiments on synthesized and real datasets show the efficacy of the proposed approach.
\end{abstract}

%%
%% The code below is generated by the tool at http://dl.acm.org/ccs.cfm.
%% Please copy and paste the code instead of the example below.
%%


%% Keywords. The author(s) should pick words that accurately describe
%% the work being presented. Separate the keywords with commas.

%%
%% This command processes the author and affiliation and title
%% information and builds the first part of the formatted document.
\maketitle

\section{Introduction}
%Humans are aware of objects that are occluded. The question is how we represent these occluded parts of perceived objects: this is the problem of amodal perception.

\begin{figure}[!t]
  \centering
  \includegraphics[width=1.0\linewidth]{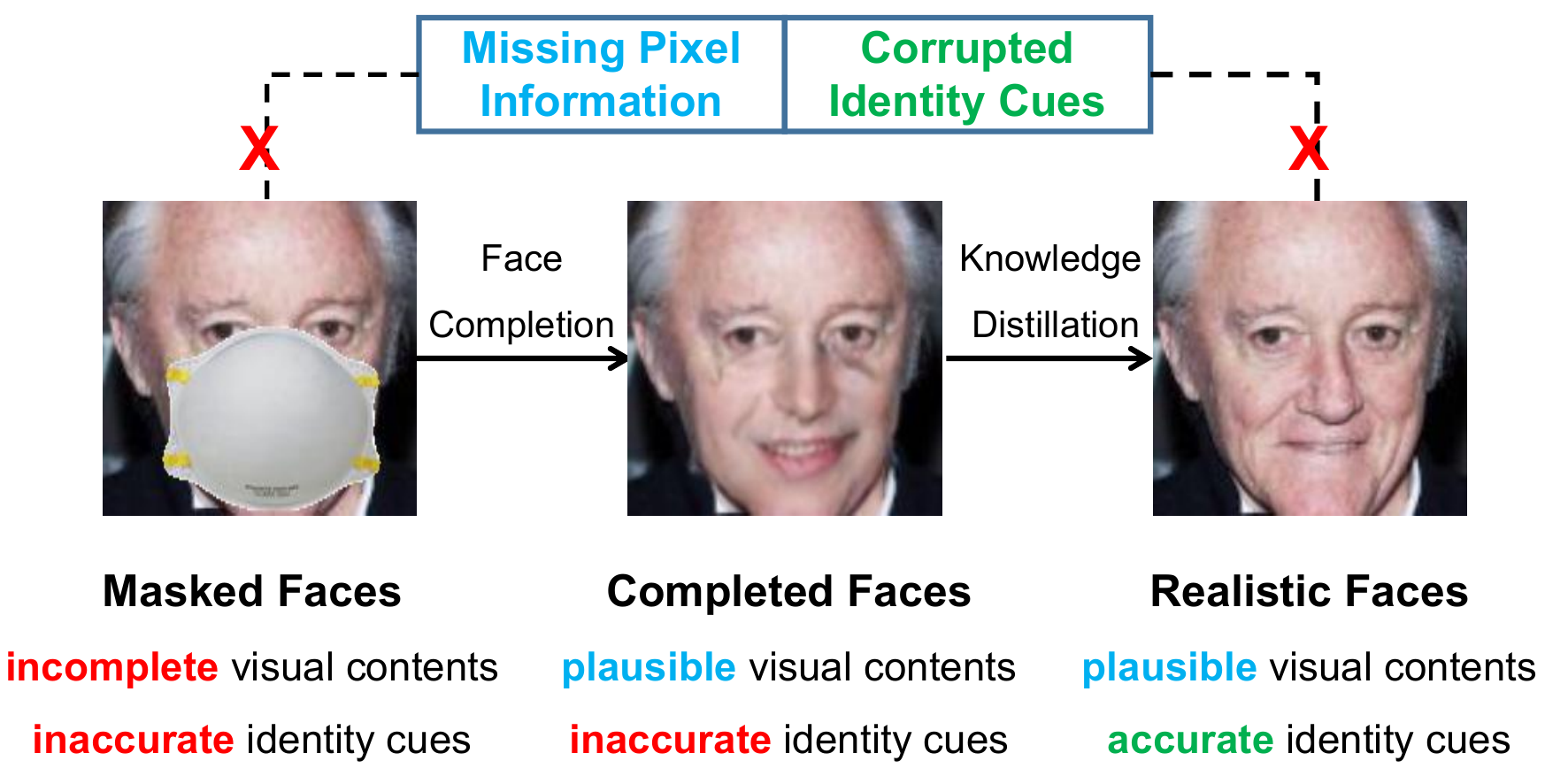}
  \caption{Inspired by the mechanism of amodal perception, we propose to solve masked face recognition via de-occlusion distillation that first enforces face completion, then inherits rich knowledge from pre-trained recognizor via distillation. In this way, both incomplete visual contents and inaccurate identity cues can be well recovered.}
  %\caption{Motivation of this work. Inspired by the mechanism of amodal perception, we propose to solve masked face recognition in two stages. We explicitly enforce face completion in the first stage. The recognizor for completed faces is then trained via relational knowledge distillation. In this way, the two challenges of incomplete visual content and insufficient identity cues can be well addressed.}
  \label{fig:motivation}
\end{figure}

Human faces in the wild are often occluded by masks, intentionally or unintentionally. The ability to handle recognition towards masked faces is essential for many visual applications,~{\em e.g.}~video surveillance~\cite{kim2005new} and urban governance~\cite{Hossain11surveillance}. In the last few years, the deep-learning-based face recognition models~\cite{taigman2014deepface,sun2014deep,schroff2015facenet,wen2016discriminative,cao2018vggface2,zhao20183d} have been able to achieve or even exceed human-level performance on public benchmarks. This has much attributed to our growing understanding of how our brain may be solving the identity recognition task. However, face recognition under more challenging conditions, such as masked faces, is less characterized. The question is how to represent these masked parts of perceived faces: this is the problem of amodal perception~\cite{Michotte1964Les}.

To facilitate recognition towards masked faces, intuitive methods~\cite{wright2008robust, yang2011robust} that seek to extract credible features from visible regions permit direct mapping between the input partial observation in the source domain and expected output in the target domain. They usually split the input image into several local parts and predict the possibilities of each being masked, on which the allocated weights for local parts are based. These approaches are supported by biological neural science conclusions that masked objects are selectively perceived as disconnected elements~\cite{kovacs1995selectivity, chen2009time}.

Recent researches have proved amodal completion indispensable in perceiving partly-observed objects. In~\cite{chen2009time} Chen~{\em et al.}~revealed that amodal completion is first manifested in face-selective areas, then implemented through feedback and recurrent processing among different cortical areas. Via attention mechanism and context restraints, many effective algorithms for occluded object recognition have been proposed~\cite{wen2016discriminative,Kortylewski2020Compositional}. However, different from general object or shape recognition, facial identity relies on both shape/view and appearance~\cite{chang2017code}. The latter involves more complexity and ambiguity when being masked. Given an example, with an image of a man with a face mask, it's rather easy to tell from isolated parts like eyes that there is a man, while difficult to tell his identity. Amodal perception for masked faces is hard to achieve by conventional Deep Neural Networks~(DNNs), where completion is done implicitly.

We propose to explicitly regularize the amodal completion process.
The most related task is face completion. Recent approaches~\cite{pathak2016context,iizuka2017globally,li2017generative,yu2018generative} based on Generative Adversarial Networks~(GANs)~\cite{goodfellow2014generative}, which formulate completion as a conditional image generation problem, have evolved as compelling tools that generate photo-realistic results. However, recently Joe~{\em et al.}~\cite{Joe2019Does} looked into the question that whether generative face completion helps recognition. Experimental results showed that, though completion greatly please biometric systems, the benefit for recognition is limited.
We suspect that it arises from the innate selectivity-invariance pattern of CNNs, where higher layers of representation amplify shared aspects and suppress irrelevant variations~\cite{Lecun2015Deep}. General constraints on efficient generative completion naturally lead to an axis representation with strong appearance bias.
There exists a clear domain gap between the inexact visual imagery~\cite{nanay2007four} and realistic faces.

In this work, inspired by the amodal completion mechanism in the human brain, we propose a novel de-occlusion distillation framework to deal with the task of masked face recognition, as shown in Fig.~\ref{fig:motivation}. The model consists of two main modules, \textit{de-occlusion} and \textit{distillation}. The de-occlusion module applies a GAN-based face completion network to eliminate the appearance ambiguity and enables the masked face to be perceived as a whole. The attention mechanism is introduced to teach the model to ``look'' at informative areas. Then to subsequently benefit recognition, the distillation module takes a pre-trained general face recognition model as the teacher and adapts its knowledge to completed faces through knowledge distillation. Recently Ge~{\em et al.}~\cite{IDGAN} employed identity-centered regularization and gained an effective accuracy boost, which inspired us to exploit in deep generative models rich problem structures and domain knowledge. Assuming the distribution of unmasked faces could provide essential guidance, we represent the teacher knowledge with structural relations among instances. Via enforcing various orders of structural similarities to provide a posterior regularization, the student learns to perform accurate recognition towards completed faces.
We evaluate the proposed method on both synthetic masked face datasets~(Celeb-A~\cite{Liu2015CelebA} and LFW~\cite{LFWTech}) and realistic masked face datasets~(AR~\cite{MaB1998AR}), both showing compelling improvements on recognition accuracy.

Our main contributions can be summarized as three folds: 1) We propose a novel end-to-end framework for masked face recognition, which first enforces face completion explicitly and then transfer rich domain knowledge from pre-trained general face recognition model via knowledge distillation; 2) We introduce the theory of amodal perception to shed light on the masked face recognition task, and our empirical results echo the theory and 3) We conduct extensive experiments to demonstrate the efficacy of our approach.

\section{Related Works}
\subsection{Amodal Perception and Face Completion}
Humans are able to recognize objects even when they are partially occluded by another pattern, so easily that one is usually not even aware of the occlusion. The phenomena of completion of partly occluded shape have been termed ``amodal perception''~\cite{Michotte1964Les}, since the occluded contours are not seen. Kovacs~{\em et al.}~\cite{kovacs1995selectivity} found that single IT units remain selective for shape outlines under a variety of partial occlusion conditions, physiologically locating where amodal perception happens for the first time. So one last question we care about is: how the occluded contents are represented. In~\cite{kovacs1995selectivity} the discrimination performance was found much better when they were familiar with the subjects. This suggests that amodal perception relies heavily on our background knowledge of how the occluded parts of the object~(may) look. They also find that the IT cells only respond to selective fragments, and conclude that amodal completion doesn't happen. Chen~{\em et al.}~\cite{chen2009time} delves into the time course of amodal completion in face perception. Their results suggest amodal completion is first manifested in face-selective areas, then implemented through feedback and recurrent processing among different cortical areas. Therefore, amodal completion plays an indispensable role in perceiving partly-observed objects.

Face completion, or inpainting, aims to recover masked or missing regions on faces with visually plausible contents. Traditional exemplar-based approaches~\cite{barnes2009patchmatch,he2014image} searched similar patches as reference for the synthesis of missing regions. While this non-parametric manner achieves good results when similar content is available, the mechanism is not scalable for objects with unique textures,~{\em e.g.}~faces. Recently, the GAN-based architecture has been widely adopted in completion with visually satisfactory results~\cite{pathak2016context,li2017generative,iizuka2017globally,Ren_2019_ICCV,Yu_2019_ICCV}. They usually train an auto-encoder to predict the missing region using a combination of reconstruction loss and adversarial loss. Despite their capacity in recovering high-quality visual patterns, the recognition accuracy gain is still limited~\cite{Joe2019Does}.

\subsection{Occluded Object Recognition}
Partial occlusions are one of the greatest challenges for many vision tasks,~{\em e.g.}~classification~\cite{Kortylewski2020Compositional}, recognition~\cite{zhao20183d}, and person re-ID~\cite{Li2019ICCV}. Various approaches have been proposed to solve the problem, following ``representation'' or ``representation'' idea. The ``representation'' idea seeks to obtain robust representations for occluded objects by decreasing or excluding the influence of missing regions and tapping the useful information. Some methods first segmented a face image into several local parts and then described the face using the ordered property of facial parts~\cite{wen2016discriminative} or extracting discriminative components~\cite{li2016robust,Kortylewski2020Compositional}. Other methods directly take the whole face image as input instead, and represent it with a good descriptor, such as sparse representation~\cite{yang2011robust} and low-rank regularization~\cite{qian2014robust}.

Different from the ``representation'' idea, the ``reconstruction'' idea utilizes the redundancy of images and performs information recovery before recognizing. Deng {\em et al.}~\cite{deng2011graph} proposed an exemplar-based Graph Laplace algorithm to complete masked faces. In this way, the approach can use the completed faces to boost the recognition accuracy. It performs well when a similar appearance and expression can be found in the library. However, the type and shape of the occlusions are innumerable and unpredictable in real scenarios, which limits its applications. More recently, GAN-based face completion approaches~\cite{pathak2016context,iizuka2017globally,li2017generative,xie2018cooperative,yu2018generative,Ren_2019_ICCV,Yu_2019_ICCV} have achieved remarkable improvement in extracting high-level contextual representations and generating photo-realistic results. However, the identity consistency during completion is less considered. ~\cite{zhang2017demeshnet,zhao20183d} enforce identity preservation through perceptual loss. To exploit structural domain knowledge,~\cite{ren2019structureflow} proposes a structural loss to constrain the structure of the generated image. Alternatively Ge~{\em et al.}~\cite{IDGAN} exploit structural domain knowledge in feature space and employ identity-centered regularization.

\subsection{Knowledge Distillation and Transfer}
Transfer learning aims to mitigate the burden of manual labeling for machine learning by transferring information between different domains or tasks. The most common approach is to fine-tune models pre-trained on public datasets like ImageNet~\cite{deng2009imagenet} for specific tasks with labeled data. Recently, as a special branch in transfer learning, knowledge distillation has gained much interest and exhibited remarkable capability in knowledge transfer. Knowledge distillation was first introduced by~\cite{buc2006model} and~\cite{hinton2015distilling} presented a more general approach within the scope of a feed-forward neural network. By using the softmax output of the teacher network as soft labels instead of hard class labels, the student model can learn how the teacher network studied given tasks in a compressed form. Romero~{\em et al.}~\cite{romero2014fitnets} improved the method by using not only the final output but also intermediate hidden layer values of the teacher network to train the student network. To encourage the diversity of learning, Luo~{\em et al.}~\cite{luo2016face} utilized the ensemble of multiple networks as the teacher to train a compact student network for face recognition. All these methods assumed that the input data of the teacher and student model are from the same domain. To boost the domain adaptation task, Su and Maji~\cite{su2016adapting} proposed cross quality distillation to learn models for recognizing low-resolution images, non-localized objects, and line-drawings by using soft labels of high-resolution images, localized objects, and color images, respectively. Radosavovic~{\em et al.}~\cite{radosavovic2018data} proposed data distillation to ensemble predictions from multiple transformations of unlabeled data to automatically generate new training annotations.

\section{De-Occlusion Distillation}
In this section, we first provide an overview of our proposed approach, then describe the details of each network component as well as the loss functions.

\begin{figure*}[!t]
  \centering
  \includegraphics[width=1.0\linewidth]{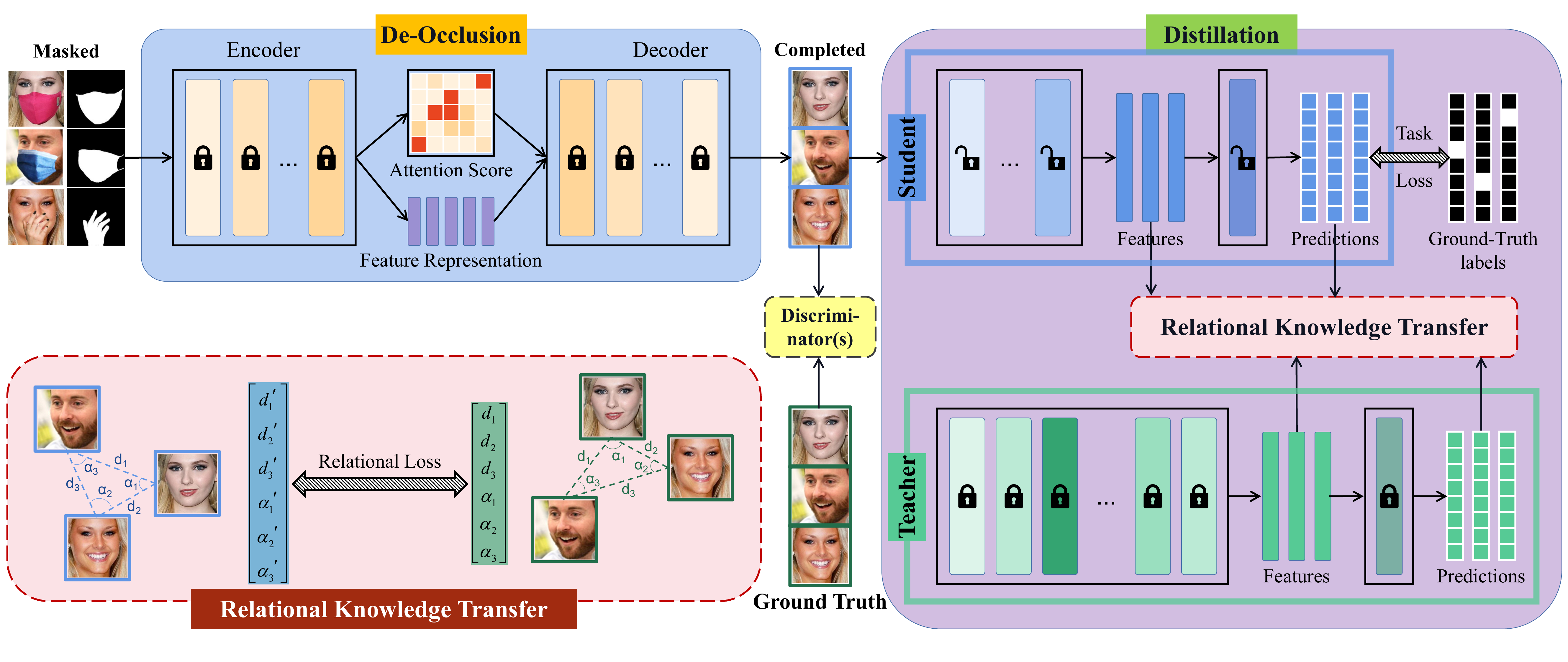}
  \caption{Overview of our proposed framework. The learning process consists of two stages. In the first stage, we initialize the input of the student model via an inpainting model. In the second stage, we do cross-quality knowledge distillation and transfer the knowledge contained in the teacher recognizor for normal faces into student recognizor by enforcing relational structure consistence. In this manner, the student network for recognizing masked faces learns representations for completed faces with the same clustering behaviors as the original ones, which could greatly benefit recognition accuracy.}
  \label{fig:architecture}
\end{figure*}

\subsection{Problem Formulation}
Masked faces are faces that not fully observed. In this section, we dissect the problem of masked face recognition and try to provide simple yet solid explanations for two questions: i) What help does data recovery do? ii) What needs to do after data recovery?

Here we describe the generative process for partly observed data, following the setting of missing data processing~\cite{Little_2020_Statistical}. Let $\textbf{X}\in\mathbb{R}^n$ be a data vector and $\textbf{M}\in\{0,1\}^n$ is a binary mask indicating which entries in $\textbf{X}$ to reveal: $x_d$ is observed if $m_d=1$ and vise versa.
\begin{equation}
  \textbf{X}\sim p_{\theta}(\textbf{X}),~~\textbf{M}\sim p_{\varphi}(\textbf{M}|\textbf{X}),
\end{equation}
where $\theta$ denotes the parameters of data distribution and $\varphi$ denotes the parameters of the mask distribution. The mask distribution is usually assumed to depend on the data $\textbf{X}$. Let $\textbf{X}_{o}$ denote the observed elements of $\textbf{X}$, and $\textbf{X}_{m}$ denote the missing elements according to the mask $\textbf{M}$. We define the target attribute as $\textbf{a}_t$. In the standard maximum likelihood setting, the unknown parameters are estimated by maximizing the following marginal likelihood, integrating over the unknown missing data values:

\begin{equation}\label{eq:int}
  \begin{aligned}
    p(\textbf{a}_t,\textbf{X}_{o}) = \int p_{\theta}(\textbf{X}_{o},\textbf{X}_{m})\cdot p_{\varphi}(\textbf{M}|\textbf{X}_{o},\textbf{X}_{m})\cdot p_{\psi}(\textbf{a}_t|\textbf{X}_{o},\textbf{X}_{m})~~d\textbf{X}_{m},
  \end{aligned}
\end{equation}
where $p_{\psi}(\textbf{a}_t|\textbf{X}_{o},\textbf{X}_{m})$ is a recognizor that gives prediction with $\textbf{X}_{m}$ replacing the missing region. Due to the ambiguity introduced by masks, the optimization process involves integration over literally infinite possible $\textbf{X}_{m}$. Even with the remarkable capacity of well-crafted DNNs, it is difficult to reach convergence.
One simple technique is to multiply with an impulse function $\delta(\textbf{X}_{m}-\hat{\textbf{X}}_{m})$:

\begin{equation}
  \delta(\textbf{X}-\textbf{X}_0)=\left\{
  \begin{aligned}
  \infty,~~~~\textbf{X} = \textbf{X}_0 \\
  0,~~~~\textbf{X} \neq \textbf{X}_0
  \end{aligned}
  \right.
\end{equation}

The physical meaning for this operation is to select the best restoration $\hat{\textbf{X}}_{m}$ for the missing data based on certain criterions,~{\em e.g.}~the coherence with the observed data regarding the wearing mask, measured by $p_{\theta}(\textbf{X}_{o},\textbf{X}_{m}) p_{\varphi}(\textbf{m}|\textbf{X}_{o},\hat{\textbf{X}}_{m})$. In this way, the optimization problem in Eq.~\ref{eq:int} can be simplified as:

\begin{equation}
  \begin{aligned}
    p(\textbf{a}_t,\textbf{X}_{o}) &=\int~~p_{\theta}(\textbf{X}_{o},\textbf{X}_{m})\cdot p_{\varphi}(\textbf{m}|\textbf{X}_{o},\textbf{X}_{m})\\
    & p_{\psi}(\textbf{a}_t|\textbf{X}_{o},\textbf{X}_{m})\cdot\delta(\textbf{X}_{m}-\hat{\textbf{X}}_{m})~~d\textbf{X}_{m}\\
    &=~~p_{\theta}(\textbf{X}_{o},\hat{\textbf{X}}_{m})\cdot p_{\varphi}(\textbf{m}|\textbf{X}_{o},\hat{\textbf{X}}_{m})\cdot p_{\psi}(\textbf{a}_t|\textbf{X}_{o},\hat{\textbf{X}}_{m}),
  \end{aligned}
\end{equation}
\begin{equation}\label{Eq:x_mis}
    \hat{\textbf{X}}_{m}~~=~~\mathop{\arg\max}_{\textbf{X}_{m}}~~p_{\theta}(\textbf{X}_{o},\textbf{X}_{m})\cdot p_{\varphi}(\textbf{m}|\textbf{X}_{o},\hat{\textbf{X}}_{m}),
\end{equation}
which turns out to be proportional to the prediction towards the completed data $p_{\psi}(\textbf{a}_t|\textbf{X}_{o},\hat{\textbf{X}}_{m})$, as the former two terms can be seen as constants once $\hat{\textbf{X}}_{m}$ is decided. Via data recovery, we turn the intricate problem in Eq.~\ref{eq:int} into two sub-problem: finding the best restoration $\hat{\textbf{X}}_{m}$ and acquiring accurate prediction $p_{\psi}(\textbf{a}_t|\textbf{X}_{o},\hat{\textbf{X}}_{m})$. This answers for our first question.

Naturally we wonder if $p_{\psi}(\textbf{a}_t|\textbf{X}_{o},\hat{\textbf{X}}_{m})$ could adopt a pre-trained state-of-the-art recognition model for faces in the wild. Following our former analysis, the optima $\hat{\textbf{X}}_{m}$ is obtained by solving a maximum optimization. Numerical solutions for these high-order space concerning optimizations are not available. In practice, we approach the optima in a data-driven way. This leads to a recovery process that is logically tractable while not accurate enough. Experiments in~\cite{Joe2019Does} support our theory. As it claims, though face completion seems pleased for a biometric system, for recognition tasks, its contribution is limited. The tractability decreases the representation capacity of the model, making it hard to deal with the turbulence in feature space, brought by varied masks. The deficiency of accuracy also suggests the domain gap regarding latent identity features. To sum up, a general recognition model is not the best fit for recognizing completed faces, yet a knowledgeable adaptation source.

In this work, we propose to solve masked face recognition with $\textbf{MFR}=\{\mathcal{G};\mathcal{R}\}$, where $\mathcal{G}$ and $\mathcal{R}$ denote a de-occlusion module and a distillation module separately. The missing information is recovered first in image space via inpainting-based de-occlusion, then in feature space via knowledge distillation. We use a de-occlusion module $\mathcal{G}(\textbf{X},\textbf{M};\mathbb{W}_{\mathcal{G}})$ to approximate the maximum optimization in Eq.~\ref{Eq:x_mis}. It takes masked faces $\textbf{X}$ as input, and aims to generate completed faces $\tilde{\textbf{X}}$ by realistically approximating the ground-true faces $\textbf{Y}$, where $\textbf{M}$ denotes the binary masks labeling the masked regions with 1 inside $\Omega$ and $\mathbb{W}_{\mathcal{G}}$ is the set of model parameters. The distillation module adopts a teacher-student scheme to distill knowledge from a pre-trained teacher network $\mathcal{R}_t(\textbf{Y};\mathbb{W}_t)$ for general faces $\textbf{Y}$ into a simpler student network $\mathcal{R}_s(\tilde{\textbf{X}};\mathbb{W}_s)$ for completed faces $\tilde{\textbf{X}}$ by transferring structural relational knowledge. Here, $\mathbb{W}_t$ and $\mathbb{W}_s$ refer to the model parameters for the teacher and student, respectively. In this way, we formulate the final goal function as:

\begin{equation}
  \begin{aligned}
    \mathop{\max}_{\mathbb{W}_{\mathcal{G}} ,\mathbb{W}_{\mathcal{R}}} \mathcal{R}(\textbf{a}_t|\textbf{X},\mathcal{G}(\textbf{X},\textbf{M};\mathbb{W}_{\mathcal{G}});\mathbb{W}_{\mathcal{R}})~~-~~\\ \mathbb{E}(\mathcal{R}_t(\textbf{Y};\mathbb{W}_t), \mathcal{R}(\textbf{X},\mathcal{G}(\textbf{X},\textbf{M};\mathbb{W}_{\mathcal{G}});\mathbb{W}_{\mathcal{R}})).
  \end{aligned}
\end{equation}

\subsection{Appearance Recovery via Completion}
In the de-occlusion module, we explicitly enforce amodal completion via a generative face completion model. First, it's important to emphasize that amodal perception is manifested in face-selective areas, and masked faces are perceived as disjointed segments. Attention plays a very important role here. We adopt the same architecture as in~\cite{yu2018generative}, which consists of a generator for inpainting and two auxiliary discriminators for regularizing from local and global views, with a contextual attention mechanism.

\noindent\textbf{Generator}~~Given an image of a masked face and a binary mask indicating the missing regions, the generator $\mathcal{G}(\textbf{X},\textbf{M};\mathbb{W}_{\mathcal{G}})$ aims to generate a photo-realistic result as similar with the ground-truth as possible. To achieve that, a pixel-wise reconstruction loss is employed to penalize the divergence, formulated as:

\begin{equation}\label{eq:Lg}
  \mathcal{L}_{\mathcal{G}}=\ell_1(\tilde{\textbf{X}}, \textbf{Y})=\ell_1(\mathcal{G}(\textbf{X},\textbf{M};\mathbb{W}_{\mathcal{G}}), \textbf{Y}).
\end{equation}

\noindent\textbf{Local and Global Discriminators}~~Two discriminate networks are adopted to identify whether input images are real or fake from global and local views, respectively. The global discriminator takes the whole image as input, while the local one uses the completed region only. Contextual information from local and global views compensate each other, eventually reaching a balance between global consistency and local details. They regularize the generator via local and global adversary losses:
\begin{equation}\label{eq:Ld}
    \begin{aligned}
      \mathcal{L}_{\mathcal{D}_i}=&\min\limits_{\mathcal{G}}\max\limits_{\mathcal{D}_i} \mathbb{E}[\log\mathcal{D}_i(\textbf{Y})+\log(1-\\
      &\mathcal{D}_i(\mathcal{G}(\textbf{X},\textbf{M});\mathbb{W}_{\mathcal{G}})],~i\in{\{g,l\}},
    \end{aligned}
\end{equation}
where $\mathcal{D}_i$ denotes the global discriminator when $i=g$ and the local discriminator when $i=l$.

\noindent\textbf{Contextual Attention}~~The contextual attention layer enables the generator to refer to features from the whole image and to learn long-distance semantic dependencies. It computes the similarity of patches centered in missing pixel $(m,n)$ and observed pixel $(p,q)$:
\begin{equation}
        s_{p,q,m,n}=\langle\frac{x_{p,q}}{\|x_{p,q}\|},\frac{\tilde{x}_{m,n}}{\|\tilde{x}_{m,n}\|}\rangle,
\end{equation}
where $x$ and $\tilde{x}$ denotes the masked and the completed face image, separately. The calculated similarities are then send through a softmax layer to obtain attention score for each pixel $s_{p,q,m,n}^*=softmax_{m,n}(\lambda\cdot s_{p,q,m,n})$, where $\lambda$ is a constant value. Finally the image contents are reconstructed by performing de-convolution on attention score. The contextual attention layer is differentiable and fully convolutional. Implementation details refer to~\cite{yu2018generative}.

\subsection{Identity Recovery via Distillation}

The last stage has recovered missing visual contents via GAN-based face completion. Experiments suggest activations responsible for amodal completion happen in the same place where cells are activated when we visualize objects with our eyes closed~\cite{Kosslyn1995Topographical}. It is easy to accept that between an actual visual stimulus and visual imagery, there exists a non-ignorable domain gap. We here raise our insight that, between the ground truth and the heuristic completion results, there also exists a non-ignorable domain gap. This is consistent with the unsatisfactory performance of generative face completion helping recognition applications. To bridge the gap, we propose to rearrange the identity features via knowledge distillation.

Knowledge distillation is a widely applied technology to transfer the knowledge from a cumbersome teacher network into a compact counterpart. To be general, the goal function for traditional knowledge distillation can be formulated as:

\begin{equation}\label{eq:L_l}
  \mathcal{L}_{\ell}=\sum_{x_i\in\textbf{X}}\ell(t_i,s_i),
\end{equation}
where $t_i$ and $s_i$ denote the feature representation produced by the teacher and student respectively, with $x_i$ as input; and $l$ denotes specific loss function adopted to penalize the differences.

Traditional distillation usually focuses on classification tasks, trained with the Cross-Entropy loss. During training, the output class distribution generated by the student is forced to be close to that of the teacher. In this way, the student could obtain better results than directly trained with class labels. The main reason may lie in that probability distribution over classes provided by the teacher's output, reveals relevance information between classes, therefore providing richer knowledge than ground truth labels.

However, the present distillation methods remain limited. Existing distillation methods usually focus on the point-wise similarity between representations of teacher and student. Previous researches~\cite{sun2014deep,parkhi2015deep,wen2016discriminative,deng2019arcface} have verified that instance relationships can help reduce the intra-class variations and enlarge the inter-class divergences in the feature space. Nevertheless this is rarely considered in distillation. We assume that what constitutes the knowledge is better presented by relations of the learned representations than individuals of those, and the structural distribution of unmasked faces could provide essential guidance for the identity feature rearrangement of completed faces. Besides, point-wise distillation methods usually require the teacher and student to share similar network architecture and close data domains. Here in the masked face recognition scenario, we seek to distill the rich knowledge about feature distribution for unmasked faces and use them to guide the rearrangement of that of completed faces. The common characteristics we seek here should be the aggregation behaviors, in other words, the instance relationships, which are more robust to network changes and domain shifts.

Let $\hat{\phi}_t(\textbf{Y};\hat{\mathbb{W}}_t)$ and $\hat{\phi}_s(\tilde{\textbf{X}};\hat{\mathbb{W}}_s)$ be the sub-networks composed by the first several layers of the teacher network $\phi_t(\textbf{Y};\mathbb{W}_t)$ and the student network $\phi_s(\tilde{\textbf{X}};\mathbb{W}_s)$, respectively, where $\textbf{Y}$ and $\tilde{\textbf{X}}$ is the corresponding input. $\hat{\phi}_t(\textbf{Y};\hat{\mathbb{W}}_t)$ is the feature extraction backend before the softmax layer for extracting the identity features of unmasked faces, while $\hat{\phi}_s(\tilde{\textbf{X}};\hat{\mathbb{W}}_s)$ denotes the layers before the embedding layer, used to extract features of masked faces. The training process of the student network can be described as transferring the relational structure of the output representation $\hat{\phi}_t(\textbf{Y};\hat{\mathbb{W}}_t)$ to $\hat{\phi}_s(\tilde{\textbf{X}};\hat{\mathbb{W}}_s)$, to improve the final recognition ability of $\phi_s(\tilde{\textbf{X}};\mathbb{W}_s)$. Let $\textbf{Y}^n$ and $\tilde{\textbf{X}}^n$ denote a set of $n$-order tuple of unmasked and completed faces respectively, $s_i=\hat{\phi}_s(\tilde{x}_i;\hat{\mathbb{W}}_s)$ is the student knowledge gained from a completed face and $t_i=\hat{\phi}_t(y_i;\hat{\mathbb{W}}_t)$ is the teacher knowledge distilled from the corresponding ground-truth. The loss function for $n$-order distillation process can be formulated as

\begin{equation}\label{eq:L_RKD}
  \mathcal{L}_{n}=\sum_{\substack{(y_1,\dots,y_n)\in\textbf{Y}^n , \\ (\tilde{x}_1,\dots,\tilde{x}_n)\in\tilde{\textbf{X}}^n}}\ell(\psi(t_1,\dots,t_n),\psi(s_1,\dots,s_n)),
\end{equation}
where $\psi$ is a relational potential function that measures relational similarity between given $n$-tuple of teacher and student models, and $l$ is a loss that penalizes structural difference based on that.

To efficiently transfer the relational knowledge, we here introduce relational loss in three orders, enforcing structural similarity in instance-wise, pair-wise and triplet-wise fashion, respectively.

\noindent\textbf{Instance-Wise Relational Loss}~~
Following the vanilla setting, we enforce instance-wise similarity via punishing the difference

\begin{equation}\label{eq:L_i}
    \mathcal{L}_i=\sum_{\substack{y_i\in\textbf{Y} , \tilde{x}_i\in\tilde{\textbf{X}}}}\ell_1(t_i,s_i),
\end{equation}
where $\ell_1$ loss is chosen instead of $\ell_2$ loss because it deals better with abnormal points. In this task, besides the gap between teacher and student domain, there also exists great variance within the student domain. Despite better convergence and more robustness, $\ell_2$ loss would bring unwanted smoothness.

\noindent\textbf{Pair-Wise Relational Loss}~~
Recent several works have used pair-wise relational distillation loss in tasks such as image classification~\cite{Liu_2019_CVPR}, image retrieval~\cite{Yu_2019_CVPR} and semantic segmentation~\cite{Liu_2019_Structured}. It is used to transfer pair-wise relations, especially pair-wise similarities in our approach, among instances. We formulate the pair-wise relational knowledge distillation loss as follows:
\begin{equation}\label{eq:L_p}
    \mathcal{L}_p=\sum_{\substack{(y_i,y_j)\in\textbf{Y}^2 , \\ (\tilde{x}_i,\tilde{x}_j)\in\tilde{\textbf{X}}^2}}\ell_{\delta}(\psi_p(t_i,t_j),\psi_p(s_i,s_j)),
\end{equation}
where $\ell_{\delta}$ is Huber loss, and $\psi_p(t_i,t_j)=\frac{1}{\mu}\|t_i-t_j\|_2$ is the pair-wise potential function which measures the Euclidean distance between the two instances $t_i$ and $t_j$ in a mini-batch space. $\mu=\frac{1}{|\textbf{Y}^2|}\sum_{(y_i,y_j)\in\textbf{Y}^2}\|t_i-t_j\|_2$ is a normalization factor, which enables relational structures transferring disregarding the difference in space dimensions between source and task field.

\noindent\textbf{Triplet-Wise Relational Loss}~~
The structure within a triplet could provide more strict regularization than that of a pair. Inspired by this, \cite{wonpyo2019rkd} propose a triplet-wise relational distillation loss:
\begin{equation}\label{eq:L_t}
    \mathcal{L}_t=\sum_{\substack{(y_i,y_j,y_k)\in\textbf{Y}^3 , \\ (\tilde{x}_i,\tilde{x}_j,\tilde{x}_k)\in\tilde{\textbf{X}}^3}}\ell_{\delta}(\psi_t(t_i,t_j,t_k),\psi_t(s_i,s_j,s_k)),
\end{equation}
where $l_{\delta}$ is Huber loss, and the corresponding triplet-wise potential function which measures the angle formed by the three instances $t_i$, $t_j$ and $t_k$ in a mini-batch space is formulated as:
\begin{equation}\label{eq:psi_triplet}
    \psi_t(t_i,t_j,t_k)=\langle\frac{t_i-t_j}{\|t_i-t_j\|_2},\frac{t_k-t_j}{\|t_k-t_j\|_2}\rangle.
\end{equation}
The triplet-wise relational loss transfers relationships of instance embedding by penalizing angular differences. Compared with the pair-wise potential function, the triplet-wise potential function measures structural similarity in a higher-order space, enabling more effective relational knowledge transferring.

\noindent\textbf{Total Loss}~~
The total loss is therefore formulated as:
\begin{equation}\label{hard}
  \mathcal{L}=\mathcal{L}_{CE}+\lambda_i\mathcal{L}_{i}+\lambda_p\mathcal{L}_p+\lambda_t\mathcal{L}_t,
\end{equation}
where $\mathcal{L}_{CE}$ is the Cross-Entropy loss between outputs of the teacher and student network, as defined in Eq.~\ref{eq:L_l} when $\ell$ adopts Cross-Entropy. $\mathcal{L}_i$, $\mathcal{L}_p$ and $\mathcal{L}_t$ are various orders of relational distillation loss defined in Eq.~\ref{eq:L_i}, Eq.~\ref{eq:L_p} and Eq.~\ref{eq:L_t}, respectively. The $\lambda_i$, $\lambda_p$ and $\lambda_t$ are weighting hyper-parameters to balance the loss terms.

\subsection{Implementation Details}
We build the de-occlusion module with a generative inpainting network using the same architecture as in~\cite{yu2018generative}. In the distillation module, we employ a pre-trained VGGFace2~\cite{cao2018vggface2} model as the teacher. It achieves a very high accuracy of 99.53\% on the LFW dataset~\cite{LFWTech} after alignment. The student network is composed of a ResNet-18~\cite{he2015resnet} model with a single embedding layer on top.

Our end-to-end network is implemented based on the deep learning library Pytorch. In the experiments, we set $\lambda_p=1.0, \lambda_t=2.0$. All models were trained with a mini-batch size of $128$. The initial learning rate is $lr=0.1$ and decreases to $0.1$ times every $24$ epochs. When sampling tuples of instances for the relational losses, we simply use all the tuples~(pairs or triplets) in the given mini-batch.

\section{Experiments}
In this section, the proposed de-occlusion distillation framework is systemically evaluated on both synthesized and realistic masked face datasets. We first introduce the experiment setting, then present experimental results on two datasets, finally we conduct ablation studies and discuss the function paradigm of the proposed method.

\begin{figure}[t]
  \centering
  \includegraphics[width=1.0\linewidth]{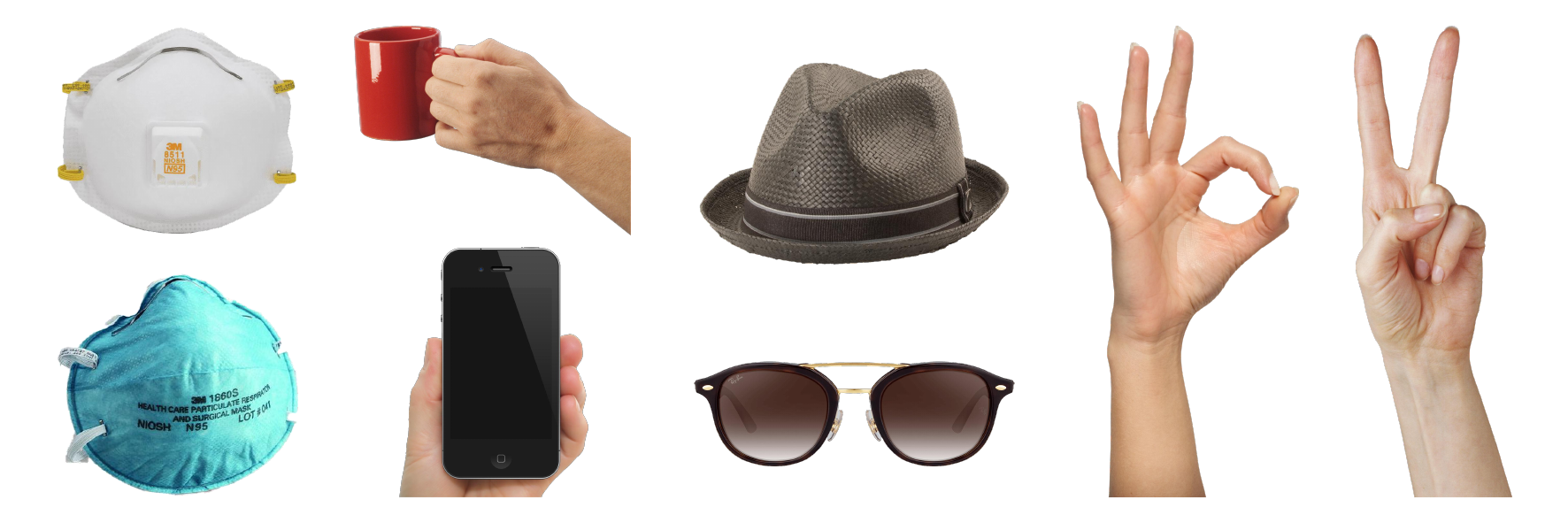}
  \caption{Examples of the masks adopted for synthetic masked faces.}
  \label{fig:masks}
\end{figure}

\subsection{Experiment Setting}

\textbf{Datasets}
Our experiments are carried out on three datasets: Celeb-A dataset~\cite{Liu2015CelebA}, LFW dataset~\cite{LFWTech} and AR dataset~\cite{MaB1998AR}.

The \textbf{Celeb-A} dataset consists of 202,599 face images covering 10,177 subjects. Each face image is cropped, roughly aligned by the position of two eyes, nose, and two mouth corners, and rescaled to $256\times256\times3$ pixels. We acquired synthetic masked faces via pasting collected masks onto these images, described later. We randomly split it into training and validation set with the ratio set as $6:1$.

The \textbf{LFW} dataset consists of 13,233 images of 5,749 identities. Same preprocessing as Celeb-A was performed to prepare the data. We used all 13,233 images on LFW to benchmark the results. $6K$ pairs~(including $3K$ positive and $3K$ negative pairs) were selected to evaluate the performance of masked face recognition.

The \textbf{AR} dataset consists of face images with varying illumination conditions, expressions, and partial occlusions. Two variations of occlusions are available in the dataset, sunglasses and scarves, which makes 1,200 images in total. We followed the same protocol to prepare the data. For our study, we randomly took an unmasked face of the same subject, instead of its non-exist original, to send into the teacher network and provide guidance.

\textbf{Synthesizing Protocols}
Considering the deficiency of masked face datasets, we synthesized masked face images by automatically pasting mask patterns into face images from Celeb-A~\cite{Liu2015CelebA} and LFW~\cite{LFWTech} Dataset.
We collected transparent mask images online. To prevent over-fitting, we followed~\cite{ge2017detecting} and divided occlusions into four categories: Simple Mask~(man-made objects with pure color), Complex Mask~(man-made objects with complex textures or logos), Human Body~(face covered by hand, hair, etc.) and Hybrid Mask~(combinations of at least two of the aforementioned mask types, or one of the aforementioned mask types with eyes occluded by glasses), and select representative masks for each type. 45 mask images are employed. Several examples of the extracted masks are shown in Fig~\ref{fig:masks}. All the masks were rescaled, covering an average of about $\frac{1}{5}$ of the face. To improve the generalization ability of the model, we did data augmentation including flipping and shift.

\subsection{Results on Synthetic Masked Faces}
\begin{figure*}
  \centering
  \includegraphics[width=1.0\linewidth]{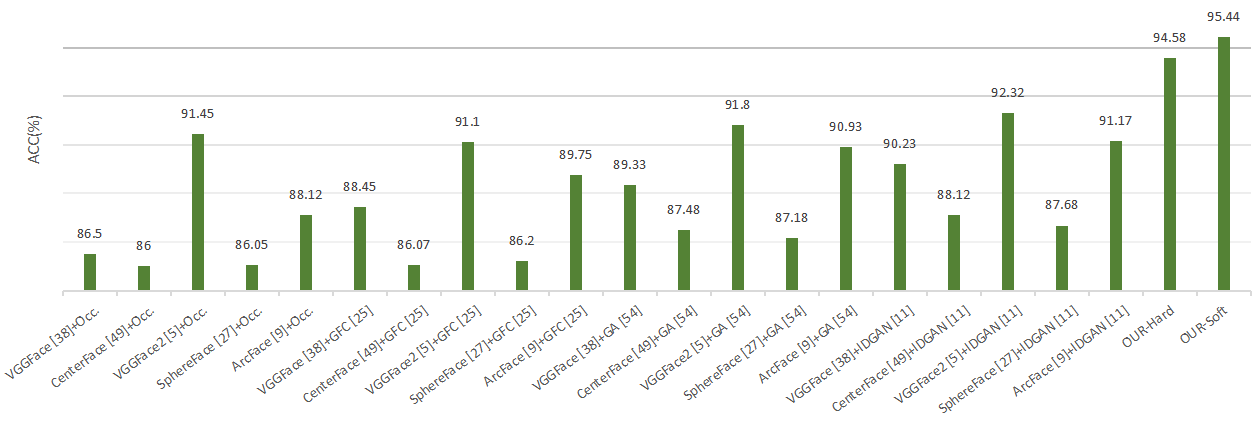}
  \caption{Evaluation accuracy of different models on LFW.}
  \label{fig:lfw_results}
\end{figure*}

\begin{table*}[t]
\small
\centering
\caption{Rank-1 recognition accuracy on AR Database by several existing recognizers. }
\begin{tabular}{cccccc|ccccc}
\hline
 & \multicolumn{5}{c}{\textbf{AR1}: Recognition of faces with glasses~(\%)} & \multicolumn{5}{c}{\textbf{AR2}: Recognition of faces with scarfs~(\%)}\\
 \hline
 & \textbf{Occ.} & \textbf{RPCA}~\cite{wright2009rpca} & \textbf{GL}~\cite{deng2011graph} & \textbf{GFC}~\cite{li2017generative} & \textbf{GA}~\cite{yu2018generative} & \textbf{Occ.} & \textbf{RPCA}~\cite{wright2009rpca} & \textbf{GL}~\cite{deng2011graph} & \textbf{GFC}~\cite{li2017generative} & \textbf{GA}~\cite{yu2018generative}\\
\hline
\textbf{PCA}~\cite{turk1991face} & 61.4 & 64.2 & 70.0 & 82.9 & 88.6 & 37.5 & 32.2 & 40.8 & 72.6 & 78.3 \\
\textbf{GPCA}~\cite{lei2008gabor} & 73.3 & 71.6 & 76.6 & 88.4 & 92.7 & 56.2 & 54.0 & 60.9 & 80.3 & 87.2 \\
\textbf{LPP}~\cite{he2003locality} & 45.7 & 61.4 & 59.0 & 83.5 & 90.0 & 43.0 & 38.3 & 47.1 & 75.9 & 80.1 \\
\textbf{SR}~\cite{wright2008robust} & 59.2 & 57.3 & 60.6 & 85.7 & 90.3 & 51.8 & 47.7 & 56.7 & 79.8 & 86.4 \\
\textbf{VGGFace}~\cite{parkhi2015deep} & 85.4 & 84.5 & 87.9 & 91.7 & 95.9 & 75.9 & 79.6 & 83.5 & 89.9 & 92.2 \\
\textbf{VGGFace2}~\cite{cao2018vggface2} & \textbf{88.3} & 86.7 & \underline{89.0} & \underline{94.2} & 97.0 & \textbf{78.2} & \underline{81.4} & \underline{85.7} & \underline{93.1} & \underline{93.3} \\
\textbf{SphereFace}~\cite{Liu2017Sphereface} & \underline{87.5} & \underline{87.2} & \underline{89.0} & 93.7 & \underline{97.5} & \underline{78.0} & 79.8 & 83.9 & 92.8 & 93.1 \\
\textbf{ArcFace}~\cite{deng2019arcface} & 85.5 & 85.2 & 87.6 & 92.3 & 95.5 & 76.4 & 79.2 & 82.6 & 90.2 & 91.9 \\
\hline
\textbf{OUR} & - & \textbf{92.1} & \textbf{93.3} & \textbf{97.2} & \textbf{98.0} & - & \textbf{84.4} & \textbf{86.8} & \textbf{93.3} & \textbf{94.1} \\
\hline
\end{tabular}
\label{tab.acc_ar}
\end{table*}

In this subsection, we compare the recognition performance for synthetic masked faces of different models. We trained our end-to-end de-occlusion model on Celeb-A, then evaluate comparison accuracy on the LFW dataset. All models extract features of all 6000 face pairs and then computes the cosine similarities between the face pairs. The accuracy is the percentage of correct prediction, where the threshold is decided as the one with the highest accuracy.

We trained our network with total loss taking the form as Eq.~\ref{hard}. Several sotas are also presented and compared. \textbf{GFC}~\cite{li2017generative}, \textbf{GA}~\cite{yu2018generative} and \textbf{IDGAN}~\cite{IDGAN} are all state-of-the-art generative inpainting methods, especially \textbf{IDGAN} is designed and optimized for masked face recognition problem. We equip them with five high-performance recognizers, and the results are shown in Fig.~\ref{fig:lfw_results}. Our model trained with Eq.~\ref{hard}, denoted as \textbf{OUR-Hard}, surpass all combinations.

During the training process, we noticed that the model converges stably at the early stages, then gets stuck soon, showing obvious over-fitting. We leave detailed illustration to Sec.~\ref{ablation} and go directly for refining strategy. With a large reduction in parameters and change on model structure, it is reasonable to suspect the student bear considerable instability, especially under perturbation settings like ours. Directly enforcing the feature to be the same could be too strict regularization. History researches indicate that soften knowledge is more efficient to learn~\cite{liu2019towards}. Therefore, we reformulate the instance-wise relational loss $\mathcal{L}_{i}$ as:
\begin{equation}\label{L*_i}
    \mathcal{L}^*_{i} = \sum_{\substack{y_i\in\textbf{Y} , \tilde{x}_i\in\tilde{\textbf{X}}}}\ell_1(t_i-\textbf{f}_{id},s_i-\textbf{f}_{id})
\end{equation}
where identity-centered feature $\textbf{f}_{id}$ represents the centroid of identity features for training images with identity label $id$:
\begin{equation}
    \textbf{f}_{id}=\frac{\sum_{i=1}^{N}\delta(y_i=id)t_i}{\sum_{i=1}^{N}\delta(y_i-id)}
\end{equation}
where $N$ denotes the size of training datasets. In experiments, the identity-centered features are pre-computed off-line. In this way, we soften the knowledge of the teacher and enable a stabler knowledge transfer. The results are shown in Fig.~\ref{fig:lfw_results} as \textbf{OUR-Soft}.

\subsection{Results on Realistic Masked Faces}
We then evaluate the proposed method on the AR dataset, where two variations of occlusions are available. For testing, the masked faces were divided into two subsets, denoted as AR1 and AR2, consisting of faces with sunglasses and scarfs, respectively.

We adopted four classic face recognition algorithms and four deep learning recognition models to test the recognition performances on the masked faces and the completed faces. Specifically, the four recognition algorithms are: 1) PCA~\cite{turk1991face}, the typical statistic-based recognition algorithm; 2) Gabor wavelet-based recognition~(GW+PCA)~\cite{lei2008gabor}, using features in the transformed domain; 3) locality projection~\cite{he2003locality}, a manifold-based recognition algorithm; and 4) SR~\cite{wright2008robust}, which is a branch of norm-based optimization. The four state-of-the-art deep learning recognition models include VGGFace~\cite{parkhi2015deep}, VGGFace2~\cite{cao2018vggface2}, SphereFace~\cite{Liu2017Sphereface} and ArcFace~\cite{deng2019arcface}.

We completed faces by two tradition methods \textbf{RPCA}~\cite{wright2009rpca} and \textbf{GL}~\cite{deng2011graph}, as well as two sota generative inpainting methods \textbf{GFC}~\cite{li2017generative} and \textbf{GA}~\cite{yu2018generative} respectively. All faces are then predicted by all recognizers above, as comparisons with our de-occlusion distillation model. The results are shown in Tab.~\ref{tab.acc_ar}, and our method achieves a higher recognition accuracy. The VGGFace2 and SphereFace model exhibit relatively milder degradation in masked scenarios among the baselines. We take this as evidence that data diversity and structural regularization are beneficial for model robustness.

\begin{figure}
  \centering
  \includegraphics[width=1.0\linewidth]{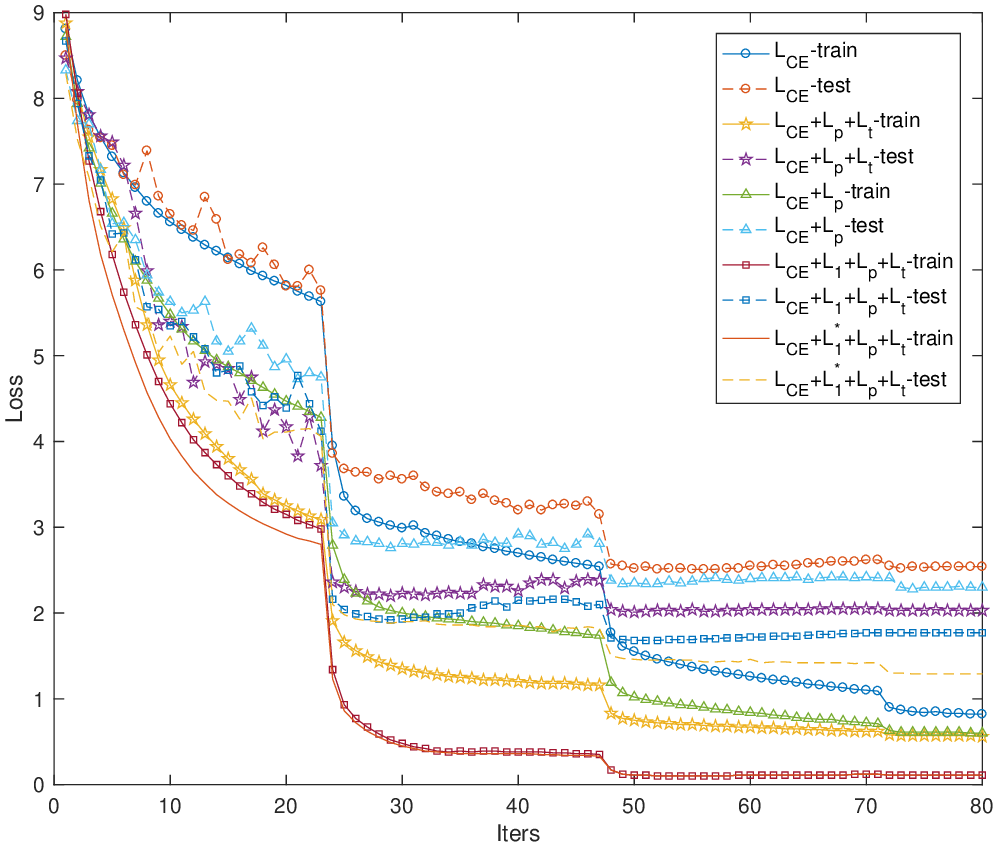}
  \caption{Loss changes with different loss settings.}
  \label{fig:ablation-loss}
\end{figure}

\subsection{Ablation Study}\label{ablation}
In this section, we conduct ablation studies to prove efficacy.

\textbf{Contribution of each loss component.} We trained our model with 1)Cross-Entropy(CE) loss only $\mathcal{L}_{CE}$, 2) CE loss and pair-wise relational loss $\mathcal{L}_{CE}+\lambda_p\mathcal{L}_p$, 3) CE loss, pair-wise and triplet-wise relational loss $\mathcal{L}_{CE}+\lambda_p\mathcal{L}_p+\lambda_t\mathcal{L}_t$, 4) CE loss, pair-wise, triplet-wise, and hard instance-wise relational loss $\mathcal{L}_{CE}+\lambda_i\mathcal{L}_{i}+\lambda_p\mathcal{L}_p+\lambda_t\mathcal{L}_t$ and 5)CE loss, pair-wise, triplet-wise, and soft instance-wise relational loss $\mathcal{L}_{CE}+\lambda_i\mathcal{L}^*_{i}+\lambda_p\mathcal{L}_p+\lambda_t\mathcal{L}_t$. Fig.~\ref{fig:ablation-loss} shows the trend of loss during training in both the training and evaluation set. Testing accuracy on Celeb-A and LFW share similar trends, and the trend on LFW is less aligned due to the domain gap. From the figure, we noticed that CE loss can well stabilize the training process. However, with CE loss only, the model fails to reach the optima. Our relational distillation loss in various orders exhibits good collaboration with CE loss and leads to the best performance eventually.

\textbf{OUR-Hard vs OUR-Soft.} It is worth to note in Fig.~\ref{fig:ablation-loss} that \textbf{OUR-Hard} trained with $\mathcal{L}_{CE}+\lambda_i\mathcal{L}_{i}+\lambda_p\mathcal{L}_p+\lambda_t\mathcal{L}_t$ shows apparent over-fitting in later stage, even worse than those trained without instance-wise loss. With a large reduction in parameters and change on model structure, it's reasonable to suspect the student bear considerable instability, especially under perturbation settings like ours. Directly enforcing the features to be exactly the same could be too strict. After we soften the instance-wise loss into identity-centered ensemble loss $\mathcal{L}^*_{i}$, as defined in Eq.~\ref{L*_i}, the resulting \textbf{OUR-Soft} shows more efficient convergence and finally reach the lowest loss. We believe this discovery is meaningful for more general adaptation tasks, especially with great domain gaps or severe perturbation.

\textbf{Choice of completion methods.} We then ask whether the choice of completion methods make a difference. We replace the face completion model in the de-occlusion module with 1)\textbf{GFC}~\cite{li2017generative}, 2)\textbf{GA}~\cite{yu2018generative} without contextual attention and 3)\textbf{IDGAN}~\cite{IDGAN}, in comparison with the adopted 4)\textbf{GA}~\cite{yu2018generative} with contextual attention. We trained model with Hard~($\mathcal{L}_{CE}+\lambda_i\mathcal{L}_{i}+\lambda_p\mathcal{L}_p+\lambda_t\mathcal{L}_t$) and Soft loss~($\mathcal{L}_{CE}+\lambda_i\mathcal{L}^*_{i}+\lambda_p\mathcal{L}_p+\lambda_t\mathcal{L}_t$) and Tab.~\ref{tab:attention} shows the results on LFW. Bold and underline denote the first and second highest in each column, separately. All models trained with Soft loss perform better than those with Hard loss, which verifies the efficacy of the softening mechanism. Besides, it's evident that the performance of \textbf{GA}~\cite{yu2018generative} with attention stands out alone, while the others are close. We attribute it to the contextual attention mechanism which allows the model to focus on the most relevant and informative areas.

\begin{table}[t]
\small
\centering
\caption{Comparisons between models adopting different face completion methods and finetuned ArcFace~\cite{deng2019arcface} (\%).}
\begin{tabular}{p{0.31\linewidth}<{\centering}p{0.04\linewidth}<{\centering}p{0.17\linewidth}<{\centering}p{0.08\linewidth}<{\centering}p{0.16\linewidth}<{\centering}}
\hline
  & \textbf{GFC} \cite{li2017generative} & \textbf{GA} \cite{yu2018generative} w/o attention & \textbf{IDGAN} \cite{IDGAN} & \textbf{GA} \cite{yu2018generative} w/ attention \\
\hline
\textbf{ArcFace}~\cite{deng2019arcface}-finetuned & 90.15 & 91.28 & 92.52 & 92.17\\
\textbf{OUR-Hard} & \underline{92.77} & \underline{92.93} & \underline{93.12} & \underline{94.58}\\
\textbf{OUR-Soft} & \textbf{93.60} & \textbf{93.55} & \textbf{93.92} & \textbf{95.44}\\
\hline
\end{tabular}
\label{tab:attention}
\end{table}

\textbf{Distillation vs Fine-tune.} Finally, to prove the efficacy of the distillation module, we fine-tune the Arcface model~\cite{deng2019arcface}, which has a similar size with our student model, on completed faces and make comparisons. The evaluation accuracy on the LFW dataset is reported in the first row of Tab.~\ref{tab:attention}. Our \textbf{OUR-Soft} model surpasses the fine-tuned ArcFace model by $3.27\%$, which suggest that fine-tuning can hardly come across the semantic gap. Our models instead, learn to recover the identity features under the guidance of pre-trained recognizer, and effectively improve the masked face recognition.

\section{Conclusion}
Masked face recognition is a problem of wide prospects for applications. Despite great efforts and advancements made over the years, current methods are restrained by incomplete visual content and insufficient identity cues. In this work, we migrate the mechanism of amodal perception and propose a novel de-occlusion distillation framework for efficient masked face recognition. The model first recovers appearance information via a generative face completion based de-occlusion module, and then transfers rich structural knowledge from a high-performance pre-trained general recognizor to train a student model.
In this way, the student model learns to recover the missing information both in appearance space and in identity space. By representing knowledge of existing high-performance recognition models with structural relations in various orders, the model is enforced to extract representations with similar aggregation behaviors with those of the teacher.
Experimental results show that the amodal completion mechanism is also beneficial for deep neural networks, and our proposed de-occlusion distillation can deal with the masked face recognition task on both synthetic and realistic datasets. In the future, we will work on the establishment of amodal perception for computer vision and further investigation on suitable network architecture.

\myPara{Acknowledgement.} This research is supported in part by grants from the National Key Research and Development Program of China (2020AAA0140001), the National Natural Science Foundation of China (61772513 \& 61922006), Beijing Natural Science Foundation (L192040) and Beijing Municipal Science and Technology Commission (Z191100007119002). Shiming Ge is also supported by the Youth Innovation Promotion Association, Chinese Academy of Sciences.

\newpage

{\small
\bibliographystyle{ACM-Reference-Format}
\bibliography{mm380}
}

\end{document}